\documentclass[conference]{IEEEtran}

\newcommand{\figref}[1]{\figurename~\ref{#1}}
\newcommand{\tabref}[1]{\tablename~\ref{#1}}
\newcommand{\eq}[1]{\eqref{#1}}

\newcommand{\ie}[0]{{i.e.,}  }
\newcommand{\eg}[0]{{e.g.,}  }

\usepackage{multicol}
\usepackage{multirow}
\newtheorem{theorem}{Theorem}
\newtheorem{definition}{Definition}
\newtheorem{Proposition}{Proposition}
\newenvironment{proof}{{\noindent\it Proof. }\quad}{\hfill $\square$\par}
\usepackage{booktabs}
\usepackage{ulem}
\usepackage{enumitem}
\usepackage{subfig}
\usepackage{color}
\usepackage{hyperref}

\usepackage{cite}
\usepackage{amsmath,amssymb,amsfonts}
\usepackage{algorithmic}
\usepackage{algorithm}
\usepackage{graphicx}
\usepackage{textcomp}
\usepackage{xcolor}
\def\BibTeX{{\rm B\kern-.05em{\sc i\kern-.025em b}\kern-.08em
    T\kern-.1667em\lower.7ex\hbox{E}\kern-.125emX}}

\begin{document}
\title{GAN Inversion for Image Editing\\
via Unsupervised Domain Adaptation
\thanks{Identify applicable funding agency here. If none, delete this.}
}

\author{\IEEEauthorblockN{Siyu Xing$^{1,4}$, Chen Gong$^{2}$, Hewei Guo$^{3}$, Xiao-yu Zhang$^{1,4,*}$, Xinwen Hou$^{3}$, Yu Liu$^{3}$}
\IEEEauthorblockA{$^1$\textit{Institute of Information Engineering, Chinese Academy of Sciences}, Beijing, China\\
$^2$ \textit{University of Virginia}, Charlattesville, The United States of America\\
$^3$\textit{Institute of Automation, Chinese Academy of Sciences}, Beijing, China\\
$^4$\textit{School of Cyber Security, University of Chinese Academy of Sciences}, Beijing, China \\
\{xingsiyu, zhangxiaoyu\}@iie.ac.cn,  chengong@virginia.edu, \{guohewei2020, xinwen.hou, yu.liu\}@ia.ac.cn
}
}

\maketitle
\begin{abstract}
Existing GAN inversion methods work brilliantly in reconstructing
high-quality (HQ) images 
while struggling with more common low-quality (LQ) inputs in practical application.
 To address this issue, we propose Unsupervised Domain Adaptation (UDA) in the inversion process, namely UDA-inversion, for effective inversion and editing of both HQ and LQ images.
Regarding unpaired HQ images as the source domain and LQ images as the unlabeled target domain, we introduce a theoretical guarantee: loss value in the target domain is upper-bounded by loss in the source domain and a novel discrepancy function measuring the difference between two domains.
Following that, we can only minimize this upper bound to obtain accurate latent codes for HQ and LQ images.
Thus, constructive representations of HQ images can be spontaneously learned and transformed into LQ images without supervision.
UDA-Inversion achieves a better PSNR of 22.14 on FFHQ dataset and performs comparably to supervised methods.
\end{abstract}

\begin{IEEEkeywords}
GAN (generative adversarial networks), GAN inversion, unsupervised domain adaptation, image editing, StyleGAN
\end{IEEEkeywords}

\section{Introduction}%
\begin{figure*}[!t]
\hspace{-3mm}
	\includegraphics[scale=0.575]{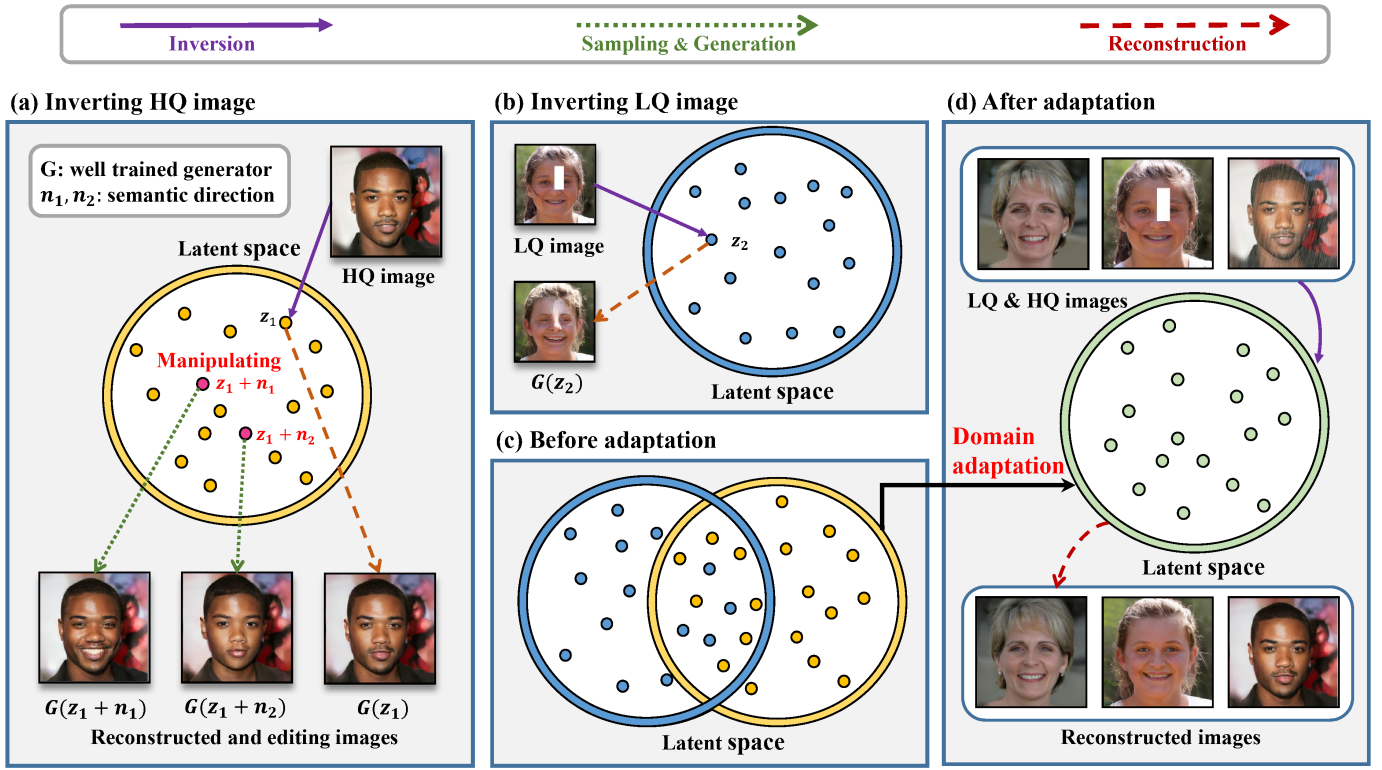}
	\caption{Overview of existing GAN inversion methods and our UDA-inversion. In subfigure (a), a naive GAN inversion maps HQ image into the latent space of a GAN model and recovers the input with latent code $\mathbf z_1$. Manipulating $\mathbf z_1$ in semantic directions can edit the reconstructed image.  (b) describes that most classical GAN inversion methods only work on HQ images, and the inverted image is usually inaccurate when inputting LQ image. 
 The image $G(\mathbf z_2)$ follows the method presented in \cite{idinvert}, a well-established encoder-based method.
 (c) illustrates that we approximate the LQ images' latent code distribution by domain adaptation. (d) shows that after adaptation, UDA-inversion obtains photorealistic results given HQ and LQ images. 
	}\label{fig1}
\end{figure*}

\label{sec:intro}
Generative adversarial networks (GANs)  have achieved remarkable performance in image generation~\cite{gan,pggan,stylegan,styleganv2}, which utilize the neural network to synthesize high-quality images in one step.
Due to the rich semantics in the latent space of a well-trained GAN (\eg StyleGAN), controllable image editing can be achieved by altering the latent code of an image generated from GAN 
\cite{ganspace,interpGAN,wu2021stylespace}.
As presented in \figref{fig1} (a), manipulating a latent code $\mathbf z_1$ can make a high-quality face image younger or happier.
GAN inversion broadens the editing scope from synthesized images to real images. 
Typically, GAN inversion learns to map a real image back to an inverted code in the latent space of a pre-trained GAN~\cite{xia2022gan}.
The inverted code accurately reconstructs the input by the generator and enables the editing of the given image in latent space.

Previous GAN inversion methods focus only on high-quality (HQ) images that are similar to the training set of GANs while ignoring other common images in practice, \ie low-quality (LQ) images~\cite{BDinvert}.
As depicted in \figref{fig1}~(a)~and~(b), employing a low-quality (LQ) image as the input for vanilla GAN inversion~\cite{idinvert} often results in a poorly inverted image, in contrast to high-quality (HQ) images. 
A gap exists between the latent code distributions of HQ and LQ images even in the same latent space.
When faced with  out of domain images (usually LQ images), this drawback impedes the progress of GAN inversion in practical applications.

This dilemma motivates GAN community to develop a GAN inversion method for inverting and editing both LQ and HQ images.
Previous works~\cite{psp,e2style,wang2023high} have been proposed to reconstruct LQ input images in a supervised manner, focusing solely on paired HQ-LQ images~\cite{psp, e2style} or degradation information~\cite{wang2023high}, such as the mask that describes the defective region in inpainting task. 
Images in real-world scenarios are typically affected by various complex and even unknown perturbations, making it difficult for researchers to gather  the corresponding paired HQ image in terms of a given LQ image. 
Thus, inverting LQ images without supervision remains a challenging problem that deserves further exploration.

To address above issues, we propose a novel GAN inversion framework based on \textbf{U}nsupervised \textbf{D}omain \textbf{A}daptation (UDA), named UDA-inversion.
The key insight is to find unbiased latent codes for HQ and LQ images by unsupervised {d}omain adaptation.
We regard HQ and LQ images as the source and unlabeled target domains, respectively.
Since the semantics in the two domains are quite similar, it is reasonable to use the source distribution to approximate the target distribution.
Constructive representation of HQ images in the source domain can be spontaneously learned and transferred to LQ images in the target domain without paired data supervising or degradation information.
As illustrated in \figref{fig1}~(c)~and~(d), UDA-inversion projects LQ and HQ images to latent space by domain adaptation, after which we can edit images by manipulating latent codes in GAN's latent space. 
In particular, we provide a theoretical guarantee that the reconstruction loss over the target domain is upper-bounded by the loss value in the source domain and the discrepancy between two domain distributions (see Theorem 1).
We reformulate the upper bound to calculate this discrepancy simply and efficiently.
By minimizing this upper bound, we can obtain the optimal latent code of images from the source and target domains (i.e., HQ and LQ images)  and steer the latent code to achieve semantic editing.
%
Experiments on image inversion and editing tasks show that our unsupervised method achieves competitive results on image inversion and editing tasks when compared to supervised methods. 

\section{Related Work}\label{sec:related-work}
\subsection{ GAN Inversion }
 GAN inversion methods can be grouped into three categories~\cite{xia2022gan}: optimization-based, encoder-based, and hybrid methods. 
The optimization-based methods use gradient descent to optimize the latent code for  reconstructing the  image~\cite{image2stylegan,multiP,DGP}. 
 However, the above methods are time-consuming in 
 inference~\cite{multiP} -- merely inverting an image requires thousands of iterations.
The encoder-based GAN inversion method, which projects images to latent space with an encoder, has presented a better performance in semantic editing \cite{idinvert} and faster inference time. 
Hybrid methods combine two methods above, using an encoder to provide a better initialization for latent code optimization~\cite{idinvert}.
Researchers pay attention to promoting the performance of GAN inversion by refining the encoder architecture~\cite{psp,e2style,hu2022style,wang2021HFGI} and focus primarily on HQ images similar to the GAN training dataset, leading to the causes of failure for LQ images. 
We show that UDA-inversion method obtains the optimal latent codes for both HQ and LQ images through unsupervised domain adaptation.

 \subsection{Image Processing with GAN Inversion }
Recent years have witnessed various GAN inversion methods applied in image processing fields, such as image colorization~\cite{multiP,DGP} and inpainting~\cite{psp,e2style}. 
%
 Most methods require training with paired data by LQ images as input and corresponding HQ images as additional supervision~\cite{psp,e2style}.
 Some optimization-based methods~\cite{multiP,DGP} require the advanced known degradation operator so that the reconstructed image after degradation closes to the LQ image. 
%
Differently, our method inverts the degraded image to an accurate latent code that can be edited to manipulate the semantics without paired images and advanced known degradation operators.

 \subsection{ Manipulating Images in GANs' Latent Space }
%
Due to the rich semantics in the latent space of StyleGAN~\cite{interpGAN,ganspace,wu2021stylespace}, diverse methods are used to manipulate latent codes for specific visual attribute editing without retraining the generator.
%
%
%
 Shen et al.~\cite{interpGAN} utilize a pre-trained classifier to identify linear semantic directions in the latent space.
 Wu et al.~\cite{wu2021stylespace} 
 discover various channel-wise style parameters from the generator to control different visual attributes.
 H{\"{a}}rk{\"{o}}nen et al.~\cite{ganspace} utilize typical unsupervised learning strategies to explore underlying semantics in the latent space of pre-trained generators. 
Overall, editing synthesized images in the latent space of GAN has achieved great success. GAN inversion extends the methods above to real image manipulation. 
We invert a real image back to a latent code and then directly control the inverted code to obtain desired inversion and editing results.

%


\begin{figure}[t]
			\centering	\includegraphics[scale=0.3]{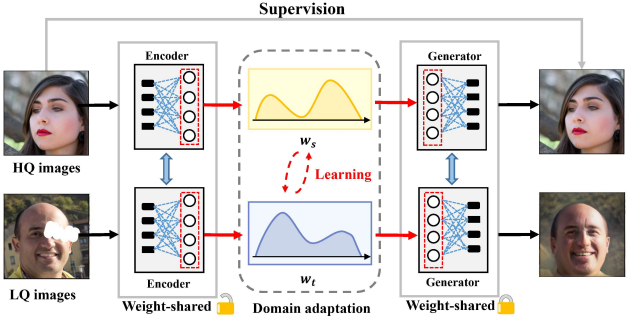}
	\caption{UDA-inversion consists of a fixed well-trained generator and a  trainable encoder, both of which are weight-shared for two domains.
  Supervision only drives inversion on HQ images.
  Latent code $\mathbf{w}_{t}$ of LQ image spontaneously learns and transfers from accurate latent code $\mathbf{w}_{s}$ of HQ image by unsupervised domain adaptation. 
 }\label{fig:framework}
\end{figure}
\section{Method}
\label{sec:bg}
In this section, we first explain problem formulation, after which we construct the total objective of  UDA-inversion.

\noindent\textbf{Problem Formulation.  }\label{subsec:pf}
We assume that $\mathcal X$ is the input data space and $\mathcal Y$ is the label space, \eg $\mathcal Y$ is a one-hot vector space for classification. 
Here, $\mathcal Y$ is an image space for GAN inversion. 
 HQ images in the source domain and LQ images in the target domain are separately drawn from the corresponding distributions $P_s$ and  $P_t$ over $\mathcal X$.
Since we have access to labeled source samples and unlabeled target samples in unsupervised domain adaptation, and it is impractical to find a HQ image corresponding to the given LQ image, HQ and LQ images  are \textit{unpaired} in UDA-inversion training.

We define $f_s$ and $f_t$: $\mathcal X\to\mathcal Y$ as the source and target labeling functions, and $\ell:\ \mathcal{Y}\times \mathcal{Y}\to\mathbb{R}^+$ is the loss function that measures the difference between the two labeling functions. Unsupervised domain adaptation intends to find a hypothesis function $h:\mathcal X\to \mathcal Y$ that generalizes to the target domain with error $    R_{t}^\ell(h)=\mathbb{E}_{\mathbf x\sim P_t}[\ell (h(\mathbf x),f_t( \mathbf x))]$ as small as possible, but we have no knowledge about $f_t(\cdot)$.


\subsection{UDA-inversion: Unsupervised Domain Adaptation in GAN Inversion}\label{sec:method}
Due to the critical scenarios above, it is impossible to directly minimize error in the target domain.
This section presents UDA-inversion to solve this dilemma.
Theorem 1 indicates the relationship between two domains, which motivates us to minimize the objective $R_{t}^l(h)$ via its upper bound.
\begin{theorem}[Generalization Bound~\cite{fdal}]\label{th-bound}
We \hspace{-0.5mm} suppose\hspace{-0.25mm} $\ell$:
$\hspace{-0.75mm}\mathcal Y\times\mathcal Y$
\hspace{-0.25mm}
$\to[0,1] $
$\subset \text{dom}~\phi^*$.
Denote $\lambda^*:=$ $ R^{\ell}_s(h^*)+ R^{\ell}_t(h^*) $, and \hspace{-0.5mm}$h^*$\hspace{-0.5mm} be the ideal joint hypothesis function under hypothesis class $\mathcal{H}$. Three terms bound the error in the target domain:
\begin{equation}\label{eq-bound}
	R_{t}^{\ell}(h) \leq R_{s}^{\ell}(h)+D_{h, \mathcal{H}}^{\phi}\left(P_{s} \| P_{t}\right)+\lambda^{*},
	\end{equation}
 where $\phi^*$ denotes  the Fenchel conjugate function in terms of a given convex function $\phi$.
 The second term of the right-hand side in \eq{eq-bound} is a discrepancy function between two distributions, formulated as:
\begin{equation}
\begin{aligned}
        	D_{h, \mathcal{H}}^{\phi}\left(P_{s}|| P_{t}\right):&= \sup _{h^{\prime} \in \mathcal{H}} \mid \mathbb{E}_{\mathbf x \sim P_{s}}\left[\ell\left(h(\mathbf  x), h^{\prime}(\mathbf  x)\right)\right] 
        \\&-
	\mathbb{E}_{\mathbf  x \sim P_{t}}\left[\phi^{*}\left(\ell\left(h(\mathbf x), h^{\prime}(\mathbf x)\right)\right)] \mid .\right.
\end{aligned}\label{eq-def2}
\end{equation}
\end{theorem}
We provide the proof of Theorem~\ref{th-bound}  and reveal the discrepancy $D_{h, \mathcal{H}}^{\phi}\left(P_{s}|| P_{t}\right)$ serving as a lower bound estimator of the $f$-divergence (e.g., KL divergence~\cite{gan,gong2021f}) { in App.A.A -A.B}.  

The last term of \eq{eq-bound} is negligible if the capability of hypothesis space $\mathcal H$ is sufficient, suggesting a convenient way to calculate objectives.
 It is common in domain adaptation that the ideal joint risk $\lambda^*$ is small  when $P_s$ $\approx$ $P_t$, which is also ubiquitous in modern GANs: 
 When latent codes $\mathbf w_1$ and $\mathbf w_2$ have a negligible difference, the generated images $G(\mathbf w_1)$ and $G(\mathbf w_2)$ are similar in visual.
Thus, the objective in target domain ${R}^{\ell}_{t}$ can be optimized by simultaneously minimizing the error in the source domain and the discrepancy between two domains in the latent space, which is formulated as:
\begin{equation}\label{eq-10}
 \mathcal{L}= \mathbb{E}_{\mathbf x \sim P_{\mathrm{s}}}[\ell( G\circ E(\mathbf x), \mathbf x)]+  \mathrm{D}_{{h}, {\mathcal{H}}}^{\phi}\left(P_{\mathrm{s}}^w\| P_{\mathrm{t}}^w\right),
\end{equation}
where  $G(\cdot)$ denotes a fixed well-trained generator and $E(\cdot)$ is an encoder.
Two domains' distributions in the latent space are defined as $P_s^w$ and $P_t^w$ with densities $p_s^w$ and $p_t^w$  respectively, and ``$\circ$'' indicates the composition operation.
As illustrated in \figref{fig:framework}, unlike supervised learning~\cite{psp,e2style} with paired images or degradation operator, UDA-inversion can minimize \eq{eq-10} to obtain accurate inversion results for both HQ and LQ images by domain adaptation.

\begin{figure*}[!t]
\centering
\includegraphics[scale=1.0]{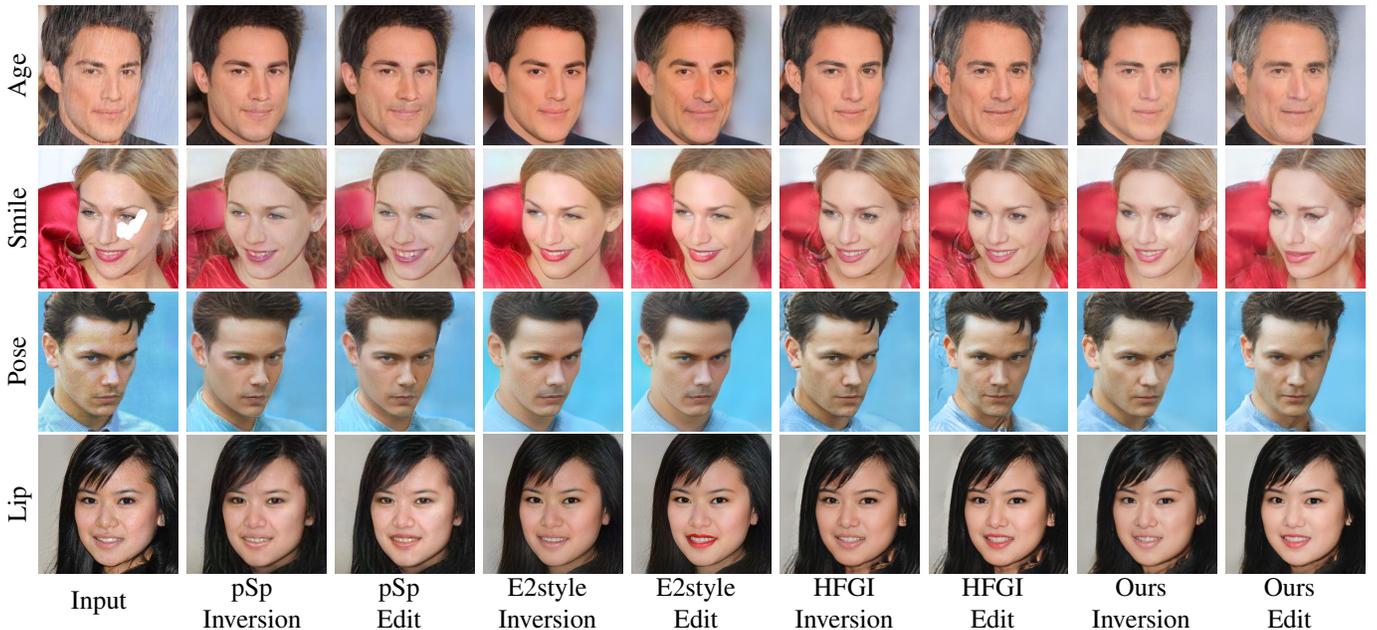}
    \caption{Qualitative results comparison on image inversion and editing with state-of-the-art GAN inversion methods. From the top to bottom, face images in the first two rows are downgraded by rain layer and random mask. Best viewed zoomed-in.}
    \label{fig6-src}
\end{figure*}

\noindent\textbf{Loss Functions in Source Domain. }\label{subsec:lfsd}
We consider the first item in \eq{eq-10} from the pixel level and other feature levels to achieve reconstruction on the source domain. 
 Specifically, we integrate pixel squared error, LPIPS (Learned Perceptual Image Patch Similarity)~\cite{LPIPS} and identity features
 to measure the difference between two images $\mathbf{x}$ and  $G\circ E(\mathbf{x})$: 
\begin{equation}
\begin{aligned}\label{eq-source-loss}
     \mathcal{L}_s&=\mathbb E_{\mathbf x\sim P_s}[\ell  (G\circ E(\mathbf x), \mathbf x)]\\
&=\lambda_1\mathcal{L}_2+\lambda_2\mathcal{L}_{LPIPS}+\lambda_3\mathcal{L}_{id}\\
     &=\mathbb{E}_{\mathbf x \sim P_{\mathrm{s}}}[\lambda_1\ell_2  (G\circ E(\mathbf x), \mathbf x)
+\lambda_2\ell_2  (H\circ G\circ E(\mathbf x), H(\mathbf x)) \\
&+\lambda_3\ell_2  (R\circ G\circ E(\mathbf x), R(\mathbf x))],
\end{aligned}
\end{equation}
where $\ell_2$ is a squared error.
 $H(\cdot)$ is the AlexNet feature extractor in LPIPS, and $R(\cdot)$ also denotes an identity extractor from a face recognition network~\cite{jiankangcvpr2019}.

\noindent\textbf{Discrepancy Between Two Domains. }\label{subsec:dbtd}
The next proposition shows how to calculate the intractable discrepancy $\mathrm{D}_{{h}, {\mathcal{H}}}^{\phi}\left(P_{\mathrm{s}}^w\| P_{\mathrm{t}}^w\right)$ between two domains in an efficient way.
\begin{Proposition} \label{prop}
Let us 
assume that
$ \forall  h\in\mathcal{ H}$, 
$\exists h'\in\mathcal{ H} $ s.t. $ \hat\ell(h'(\mathbf{w}),h(\mathbf{w}))=\phi'(\frac{p_\mathrm{s}^w(\mathbf{w})}{p_\mathrm{t}^w(\mathbf{w})}),\text{ for all latent code } \mathbf{w}\in\text{supp}(p_\mathrm{t}^w(\mathbf{w})) $.
We denotes  $d_{s,t}$ as the lower bound of the discrepancy in \eq{eq-defdst}.
Maximizing  $d_{s,t}$ can obtain equivalent  $\mathrm{D}_{{h}, {\mathcal{H}}}^{\phi}\left(P_{\mathrm{s}}^w\| P_{\mathrm{t}}^w\right)$, 
\ie $ \max_{h'\in\mathcal H}d_{s,t}=\mathrm{D}_{{h}, {\mathcal{H}}}^{\phi}\left(P_{\mathrm{s}}^w\| P_{\mathrm{t}}^w\right) $.
\begin{equation}\label{eq-defdst}
\begin{aligned}
 d_{s, t}:&=\mathbb{E}_{\mathbf w\sim P_{\mathrm{s}}^w}\left[\hat{\ell}\left({h}^{\prime}(\mathbf{w}), {h}(\mathbf{w})\right)\right]\\
 &-\mathbb{E}_{\mathbf{w} \sim P_{\mathrm{t}}^w}\left[(\phi^{*} \circ \hat{\ell})\left({h}^{\prime}(\mathbf{w}), {h}(\mathbf{w})\right)\right].
\end{aligned}
\end{equation}
\end{Proposition}
Here, $h$ and $h^\prime $ belong to
hypothesis space $\mathcal{H}$,
 $\phi^*$ is the Fenchel conjugate function of 
$\mathrm{D}_{{h}, {\mathcal{H}}}^{\phi}\left(P_{\mathrm{s}}^w\| P_{\mathrm{t}}^w\right)$,
and 
$ \hat\ell( h'(\mathbf{w}), h(\mathbf{w}))\to \text{dom}~\phi^*$
indicates the ``distance" between two functions. 
The proof of Proposition~\ref{prop} is provided in App.A.C.
In short, we approximate the intractable discrepancy by solving an  optimization problem over $h'$:
\begin{equation}\label{eq-minmax2}
\begin{aligned}
\mathrm{D}_{{h}, {\mathcal{H}}}^{\phi}\left(P_{\mathrm{s}}^w\| P_{\mathrm{t}}^w\right)&= \max_{h'\in\mathcal H}
\mathbb{E}_{\mathbf{w} \sim P_{\mathrm{s}}^w}\left[{\hat\ell}\left({h}^{\prime}, {h} \right)\right]  
\\&-\mathbb{E}_{\mathbf{w} \sim P_{\mathrm{t}}^w}\left[(\phi^{*} \circ {\hat\ell})\left({h}^{\prime} , {h} \right)\right] .
\end{aligned}
\end{equation}

To facilitate $\hat{\ell}$ and $h$ to measure  the similarity between two images, we project inverted images $G\circ E(\mathbf x)$ into the representation space of LPIPS \cite{LPIPS} to obtain the perceptual similarity between two images.
When the generator $G$ is well-trained and fixed, the discrepancy between two distributions in the latent space is equivalent to the synthesized image space. Thus, according to \eq{eq-minmax2}, the discrepancy is reformulated as:
\begin{equation}\label{eq-h-adv}
\begin{aligned}
\max _{\hat{H}} d_{s,t}&=\max _{\hat{H}}  \mathbb{E}_{\mathbf x \sim P_{\mathrm{s}}}\left[\hat{\ell}\left(\hat{H} \circ G\circ E, H\circ {G} \circ E\right)\right] \\&-\mathbb{E}_{\mathbf x \sim P_{\mathrm{t}}}\left[\left(\phi^{*} \circ \hat{\ell}\right)\left(\hat{H} \circ G\circ E, H\circ{G} \circ E\right)\right],
\end{aligned}
\end{equation}
where $\hat{H}$ indicates a network with the same structure as AlexNet
$H$ in \eq{eq-source-loss} and $\hat{\ell}$ refers to LPIPS loss, interpreted  to minimize the distance at perceptual feature level. 
Following proposition 1, we choose the convex function $\phi$ of Pearson $\chi^2$ divergence.
Detailed discussion is present in  App.A.D.

Overall, the objective in \eq{eq-10} is reformulated as:
\begin{equation}\label{eq-totalloss}
        \begin{aligned}
&\min _{ E } \max _{{\hat{H}}} \mathcal{L}_s+\lambda_{uda}d_{s,t},\\ 
\end{aligned}
\end{equation}
where $\lambda_{uda}$ is a hyper-parameter.
To optimize the min-max objective in~\eq{eq-totalloss}, we alternatively update the parameters of $E$ and $\hat H$ to relieve the unstable gradient burden.

\section{Experiments}\label{sec:exper-setup}

 \subsection{Experiment Setting}
\noindent\textbf{Datasets. }
We conduct experiments on FFHQ~\cite{stylegan} dataset for training and CelebA-HQ~\cite{pggan} test set for evaluation. 
After assigning the first 50\% of images from the training set as data in the source domain (i.e., HQ images), we degrade remaining data as the LQ images in target domain. 
 The degradation operations contain rain layer, random mask, and down-sample, some of which are utilized in existing methods~\cite{e2style,psp}. 
We use the ``RainLayer’’ feature from the imgaug library~\cite{imgaug} with default parameters, perturbing images to a rain effect.

\noindent\textbf{Implementation Details. }
We want to verify that UDA-inversion matches the performance of the state-of-the-art supervised encoder-based GAN inversion methods, i.e., pSp~\cite{psp}, E2style \cite{e2style}, HFGI \cite{wang2021HFGI}, which are selected as our baselines.
To ensure a fair comparison, we retrain these methods with the same source domain as ours and provide additional paired LQ-HQ images from the target domain. 
We use the pre-trained StyleGAN2~\cite{styleganv2} as the generator.
 We follow existing encoder-based methods ~\cite{wang2021HFGI,e2style} with Ranger optimizer~\cite{liu2019variance,zhang2019lookahead}  during training and set the initial learning rate to 0.0001.
The encoder architecture and hyper-parameters are the same as  HFGI~\cite{wang2021HFGI}.
Two popular editing methods, GANSpace~\cite{ganspace} and InterfaceGAN~\cite{interpGAN} are selected for semantic editing  to manipulate inverted images.

\noindent\textbf{Evaluation Metrics}
We quantitatively evaluate the inversion results from two aspects, i.e., inversion accuracy and image quality~\cite{xia2022gan}. 
Inversion accuracy is usually measured by PSNR (Peak Signal-to-Noise Ratio), SSIM (Structural Similarity)~\cite{ssim}, and MSE (Mean Squared Error).
Images generated based on GANs are usually evaluated by  FID (Fr\'echet inception distance)~\cite{fid} to assess the quality and diversity of image distribution.
We introduce IDs (Identity similarity)~\cite{huang2020curricularface} to evaluate the identity  consistency of editing results.

\begin{table}[!t]
\caption{Quantitative comparison of different GAN inversion methods on CelebA-HQ dataset. In the first column, `src' means the evaluation from the source domain, and `trg' means the evaluation  from the target domain. 
 $\uparrow$ means higher is better, and $\downarrow$ means lower is better.
}\label{tab-evaluation1}
\centering
\resizebox{0.85\columnwidth}{!}{
\begin{tabular}{@{}l|cccc}
\toprule
 &
  {PSNR$\uparrow$} &
  {SSIM$\uparrow$} &
  {FID$\downarrow$} &
    {MSE$\downarrow$}\\ \midrule
HFGI (src) &
  {21.97} &
  \textbf{0.64} &
  \textbf{16.97} &
  0.028  \\
HFGI (trg)    & 21.80 & 0.63  & 17.56 & 0.029  \\
E2style (src) & 21.12 & 0.62  & 47.90 & 0.035\\
E2style (trg) & 21.06 & 0.62 & 48.34 & 0.035  \\
pSp (src) & 20.37 & 0.56 & 46.84 &0.040   \\
pSp (trg) & 20.32 & 0.56 & 46.56 & 0.040 \\ \midrule
Ours (src)    & \textbf{22.14} & \textbf{0.64} &  20.09 & \textbf{0.026}  \\
Ours (trg)    & 20.22 & 0.62  & 23.23 & 0.050 \\ \bottomrule
\end{tabular}}
\end{table}

\begin{table}[!t]\caption{Quantitative comparison of editing results for FID and IDs (Identity similarity). 
\textbf{Bold} indicates the best result. \underline{Underline} indicates the second-best.}
\label{tab:editing}
\centering
\resizebox{\columnwidth}{!}{%
\begin{tabular}{@{}ccrrrrrr@{}}
\toprule
                     &         & \multicolumn{2}{c}{Age}                          & \multicolumn{2}{c}{Pose}                         & \multicolumn{2}{c}{Smile}                        \\ \midrule
                     & Methods & \multicolumn{1}{c}{FID$\downarrow$} & \multicolumn{1}{c}{IDs$\uparrow$} & \multicolumn{1}{c}{FID$\downarrow$} & \multicolumn{1}{c}{IDs$\uparrow$} & \multicolumn{1}{c}{FID$\downarrow$} & \multicolumn{1}{c}{IDs$\uparrow$} \\ \midrule
\multirow{3}{*}{Src} & HFGI    & {\underline{ 57.66}}             & \textbf{0.45}          & \underline{ 67.12}             & \underline{ 0.46}             & \underline{ 61.64}             & \underline{ 0.45}             \\
                     & E2style & 106.51                  & 0.33                   & 91.84                   & 0.39                   & 84.08                   & 0.37                   \\
                     & pSp     & 104.61                  & 0.28                   & 70.21                   & 0.25                   & 70.07                   & 0.32                   \\
                     & Ours    & \textbf{57.06}          & \textbf{0.45}          & \textbf{62.15}          & \textbf{0.47}          & \textbf{61.19}          & \textbf{0.46}          \\ \midrule
\multirow{3}{*}{Trg} & HFGI    & \textbf{59.53}          & \textbf{0.41}          & \underline{ 69.89}             & \textbf{0.42}          & \textbf{63.40}          & \textbf{0.41}          \\
                     & E2style & 108.68                  & 0.31                   & 94.42                   & 0.37                   & 85.48                   & 0.36                   \\
                     & pSp     & 104.53                  & 0.17                   & 70.29                   & 0.24                   & 70.44                   & 0.31                   \\
                     & Ours    & \underline{ 60.59}             & \underline{ 0.40}             & \textbf{68.60}          & \textbf{0.42}          & \underline{ 64.82}             & \textbf{0.41}          \\ \bottomrule
\end{tabular}%
}
\end{table}

\subsection{Results and Analysis on Inversion and Editing Tasks}\label{sec:exper-result}

\noindent\textbf{Qualitative Evaluation. }
Figure~\ref{fig6-src}  displays a visualization of the inversion and editing results.
The performance of UDA-inversion is comparable to existing supervised methods. 

\noindent {$\bullet$\textit{Inversion results. }}
The even columns of~\figref{fig6-src} present a qualitative comparison of inverted images.
Specifically, pSp~\cite{psp} and E2style~\cite{e2style} may struggle with reconstructing details, but UDA-inversion can better reconstruct finer details (\ie 1st, and 3rd rows). Compared to HFGI~\cite{wang2021HFGI}, UDA-inversion better overcomes the artifacts in the background.

\noindent {$\bullet$\textit{Editing results. }}
UDA-inversion achieves performance comparable to existing methods.
The visual alteration of E2style's editing varies slightly (\eg 2nd row), and the background of HFGI editing is distorted (\eg 2nd row and 3rd row). Our method presents the best visually pleasing editing results among baselines, suggesting that existing supervised  methods concentrate on reconstruction, while UDA-inversion generates semantically relevant latent codes through domain adaptation.%

\noindent\textbf{Quantitative  Evaluation. }
 Quantitative results on inversion and editing tasks are presented in \tabref{tab-evaluation1} and \tabref{tab:editing}.
\noindent {$\bullet$\textit{Inversion results. }}
 Specifically, UDA-inversion outperforms baselines on PSNR, SSIM, and MSE and achieves comparable results on FID for images in the source domain. 
 The loss function of UDA-inversion on the source domain is similar to baselines, indicating the positive impact of our domain adaptation on source image inversion.

\noindent {$\bullet$\textit{Editing results. }}
  We evaluate editing results  using FID and IDs (Identity similarity~\cite{huang2020curricularface}).
\tabref{tab:editing} shows a quantitative comparison 
 under the same editing scalar in each attribute.
 It is noticed that our editing results surpass other baselines in the source domain and achieve competitive results in the target domain, indicating that our method can enhance the ability to understand semantics in GANs' latent space.
\figref{fig6-src} show consistent results with quantitative evaluation. 

\subsection{Additional Analysis}
\noindent\textbf{Ablation Study.  }
\begin{figure}[!t]
         \centering
    \includegraphics[scale=0.9]{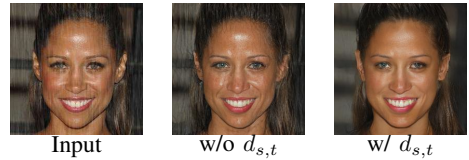}

    \caption{ Inversion result of UDA-inversion w/ and w/o $d_{s,t}$.}
    \label{fig:ablation}
\end{figure}
\begin{table}[!t]
\caption{Ablation study of UDA-inversion on target domain, evaluated by PSNR, SSIM, FID, and MSE.}
\label{tab:ablation}
\resizebox{\columnwidth}{!}{%
\begin{tabular}{@{}lcccc@{}}
\toprule
                            & PSNR$\uparrow$  & SSIM$\uparrow$ & FID$\downarrow$   & MSE $\downarrow$ \\ \midrule
UDA-inversion               & \textbf{20.22} & \textbf{0.62} & \textbf{23.23} & \textbf{0.05} \\
UDA-inversion w/o $d_{s,t}$ & 19.01 & 0.59 & 31.10 & 0.06 \\ \bottomrule
\end{tabular}%
}
\end{table}
We aim to verify the effectiveness of our major component, i.e., $d_{s,t}$ in Eq. (\ref{eq-totalloss}). As shown in \tabref{tab:ablation} and \figref{fig:ablation}, we examine the corresponding results by  comparing two inversion versions: with and without $d_{s,t}$ (denoted by w/ $d_{s,t}$ and w/o $d_{s,t}$, respectively).
UDA-inversion alleviates the negative influence of inverted images from LQ images with $d_{s,t}$, suggesting that domain adaptation makes the difference between two domains smaller, and our proposed method is more robust to perturbation.

\begin{table}[!t]
\centering
\caption{Qualitative comparison of inversion quality, run time, and model parameters on the CelebA-HQ dataset.}
\label{tab:a+b}
\centering\resizebox{1\linewidth}{!}{
\begin{tabular}{@{}lcccc@{}}
\toprule
              & PSNR$\uparrow$  & SSIM $\uparrow$ & RunTime(s)$\downarrow$ & Param(M) $\downarrow$\\ \midrule
DFDNet~\cite{Li_2020_ECCV}+HFGI~\cite{wang2021HFGI}   & 19.92 & 0.49 &     1.05s       & 395.7    \\
GPEN~\cite{YangRX021}+HFGI~\cite{wang2021HFGI}   & 19.32 & 0.57 &     0.54s       & 288.6    \\
UDA-inversion & \textbf{20.22} & \textbf{0.62} &   \textbf{0.24s}         & \textbf{262.4}    \\ \bottomrule
\end{tabular}}
\end{table}

\noindent\textbf{Additional Experiment. }
Table \ref{tab:a+b} reports a qualitative comparison between UDA-inversion and a combination of image reconstruction and GAN inversion, where the former is a blind face restoration method, i.e., DFDNet and GPEN~\cite{Li_2020_ECCV,YangRX021}. 
This comparison highlights the distinction between UDA-inversion and a straightforward ``A+B'' pipeline. 
By measuring PSNR, SSIM, average run time, and model parameters on the CelebA-HQ test set in \tabref{tab:a+b}, 
 UDA-inversion outperforms this combination in terms of inversion results with faster inference speed and fewer parameters. 

\section{Conclusions}\label{sec:conc}
This paper proposes UDA-inversion, a novel encoder-based approach for low-quality GAN inversion without the supervision of paired images or degradation information. 
We explore how to map HQ and LQ images into the latent space of GAN and derive the underlying knowledge across two distributions by unsupervised domain adaptation. 
Resolving this problem from domain adaptation, we first regard HQ and LQ images as the source and target domains.
Then, we optimize the encoder from the generalization bound of domain adaptation by minimizing the source error and the discrepancy between two distributions in the latent space.  
By effectively learning and transferring representations from HQ to LQ images, UDA-inversion shows promising results, outperforming some supervised methods. 
 This study offers a unique inspiration for latent embedding distributions in image processing tasks.

\section*{Acknowledgment}
We would like to thank Haichao Shi and Yaru Zhang for their insightful discussions during the course of this project.
This work was supported by the National Natural Science Foundation of China (NSFC) (Grant 62376265).

\normalem
\bibliographystyle{IEEEbib}
\bibliography{bibfile}
\newpage
\clearpage

\begin{center}
 \textbf{\large Appendix of GAN Inversion for Image Editing  \\  via Unsupervised Domain Adaptation }
\end{center}

\begin{appendices}

\section{Details of Proof}\label{appendix:sec:proof}
In this section, we provide the proofs for the mentioned theorem and equations in the main article.

\subsection{The lower bound of \texorpdfstring{$f$} --divergence.}\label{appendix:bound}
First, we recall the definition of $f$-divergence, from which we  exploit a discrepancy to measure the difference between two distributions.
\begin{definition}[$f$-divergence\cite{ali1966general}]\label{f-div}
The $f$-divergence $D_{\phi}$  measures the difference between two distributions $P_s$ and $P_t$ with densities $p_s$ and $p_t$, where $p_s$ is absolutely continuous with respect  to $p_t$. The function $\phi$: $\mathbb  R_{+}\to\mathbb{R}$  must  be a convex, lower semi-continuous function and satisfy $\phi(1)=0$. The $f$-divergence is formulated as:
\begin{equation}
	D_{\phi}(P_s||P_t)=\int p_t(x)\phi\left(\frac{p_s(x)}{p_t(x)}\right)dx .
\end{equation}
\end{definition}
Many popular divergences used in machine learning are special cases of $f$-divergence. For instance, the KL-divergence ${D}_{\text{KL}}(P\|Q) := \mathbb{E}_{x \sim P}[\log(p(x)/q(x))]$ is the most commonly used one \cite{gan,gong2021f,ZHANG2021383}. \tabref{tb-fdiv} lists popular divergences and convex functions.
However, $f$-divergence can not be estimated for an arbitrary distribution from finite samples. Acuna et al.~\cite{fdal} propose a method to approximate $f$-divergence. 
\begin{definition}[$D_{h,\mathcal{H}}^{\phi}$ discrepancy\cite{fdal}]
Under the same conditions of the convex function as Definition~\ref{f-div}, let $\phi^*$ be the Fenchel conjugate function of $\phi$, which is defined as $\phi^*(y):=\sup_{x\in\mathbb{R}^+}\{xy-\phi(x)\}$. The discrepancy between $P_s$ and $P_t$ can be formulated as:
\begin{equation}
\begin{aligned}
        	D_{h, \mathcal{H}}^{\phi}\left(P_{s}|| P_{t}\right):&= \sup _{h^{\prime} \in \mathcal{H}} \mid \mathbb{E}_{x \sim P_{s}}\left[\ell\left(h(x), h^{\prime}(x)\right)\right] 
        \\&-
	\mathbb{E}_{x \sim P_{t}}\left[\phi^{*}\left(\ell\left(h(x), h^{\prime}(x)\right)\right)] \mid .\right.
\end{aligned}\label{app:eq-def2}
\end{equation}
\end{definition} 
In the above equation, $\mathcal H$ is the hypothesis class.
For any $h$ and $h'\in\mathcal{H}$,  $f$-divergence and  discrepancy $D_{h,\mathcal{H}}^{\phi}$ satisfy the following inequality:
\begin{equation}\label{eq-f-lowbound}
    D_{\phi}\left(P_{s} \| P_{t}\right)\geq D_{h,\mathcal{H}}^{\phi}\left(P_{s} \| P_{t}\right).
\end{equation}
\begin{proof}
\begin{align}
    D_{h,\mathcal{H}}^{\phi}\left(P_{s} \| P_{t}\right)&= \sup _{h^{\prime} \in \mathcal{H}} \mid \mathbb{E}_{x \sim P_{s}}\left[\ell\left(h(x), h^{\prime}(x)\right)\right]
    \\&-
	\mathbb{E}_{x \sim P_{t}}\left[\phi^{*}\left(\ell\left(h(x), h^{\prime}(x)\right)\right)] \mid \right.\\
	  & \ge |\mathbb{E}_{x \sim P_{s}}\left[\ell\left(h(x), h^{\prime}(x)\right)\right]
   \\&-
	\mathbb{E}_{x \sim P_{t}}\left[\phi^{*}\left(\ell\left(h(x), h^{\prime}(x)\right)\right)]| \right.\\
	&=|R_{s}^{\ell}\left(h, h^{\prime}\right)-R_{t}^{\phi^{*} \circ \ell}\left(h, h^{\prime}\right)|.\label{proof-eq0}
\end{align}
$f$-divergences have the low bound \cite{xuanlong}:
\begin{equation}
    D_{\phi}\left(P_{\mathrm{s}} \| P_{\mathrm{t}}\right) \geq \sup _{T \in \mathcal{T}} \mathbb{E}_{x \sim P_{\mathrm{s}}}[T(x)]-\mathbb{E}_{x \sim P_{\mathrm{t}}}\left[\phi^{*}(T(x))\right].\label{eq-f-bound}
\end{equation}
Equality in \eq{eq-f-bound} holds if $\mathcal{T}$ is measurable function set \cite{fdal}.
$\mathcal T$ was restricted by the definition of $D_{h,\mathcal{H}}^{\phi}\left(P_{s} \| P_{t}\right)$, thus $ D_{\phi}\left(P_{s} \| P_{t}\right)\geq D_{h,\mathcal{H}}^{\phi}\left(P_{s} \| P_{t}\right)$.
\end{proof}

\subsection{Theorem 1.}\label{appendix:th1}
We suppose $\ell:\mathcal Y\times \mathcal Y\to [0,1]\subset \text{dom}~\phi^*$. Denote $ \lambda^*:= R^{\ell}_s(h^*)+ R^{\ell}_t(h^*) $, and $ h^* $ is the ideal joint hypothesis. The error in the target domain is bounded by three terms:
\begin{equation}
	R_{t}^{\ell}(h) \leq R_{s}^{\ell}(h)+D_{h, \mathcal{H}}^{\phi}\left(P_{s} \| P_{t}\right)+\lambda^{*} .
	\end{equation}\label{ap-bound}
	\begin{proof}
From the definition of Fenchel conjugate, we have \begin{equation}\label{proof-th1-eq1}
    \phi^*(t)=\sup_{x\in\text{dom}\phi}(xt-\phi(x))\geq t-\phi(1)=t.
\end{equation}

In this paper, $R^{\ell}_{t}(h,f_t)=\ell(f_t(\mathbf x),G\circ E(\mathbf x))=\mathbb{E}_{\mathbf x\sim p_t}||H(f_t(\mathbf x))-H(G\circ E(\mathbf x))||_2$ satisfies  triangle inequality.
This allows us to derive
\begin{align*}
    R^{\ell}_{t}(h,f_t)&\leq R^{\ell}_{t}(h,h^*)+R^{\ell}_{t}(h^*,f_t) \text{\hspace{4mm} triangle inequality}\\
    &=R^{\ell}_{t}(h,h^*)+R^{\ell}_{t}(h^*,f_t)\\
    &-R^{\ell}_{s}(h,h^*)+R^{\ell}_{s}(h,h^*)\\
    &\leq R^{\phi^*\circ\ell}_{t}(h,h^*)-R^{\ell}_{s}(h,h^*)\\
    &+R^{\ell}_{s}(h,h^*)+R^{\ell}_{t}(h^*,f_t) \hspace{2cm}\text{\eq{proof-th1-eq1}}  \\
    &\leq |R^{\phi^*\circ\ell}_{t}(h,h^*)-R^{\ell}_{s}(h,h^*)|\\
    &+R^{\ell}_{s}(h,h^*)+R^{\ell}_{t}(h^*,f_t)\\
    &\leq D_{h,\mathcal H}^{\phi}(P_s||P_t)+R^{\ell}_{s}(h,h^*)\\
    &+R^{\ell}_{t}(h^*,f_t)  \text{\hspace{4cm}\eq{proof-eq0}}   \\
    &\leq D_{h,\mathcal H}^{\phi}(P_s||P_t)\\
    &+R^{\ell}_{s}(h,f_s)+\underbrace{R^{\ell}_{s}(h^*,f_s)+R^{\ell}_{t}(h^*,f_t)}_{\lambda^*}.
\end{align*}

Furthermore, if $h^*$ is the ideal hypothesis, then $\lambda^*$ is expected to be  a rather small value. 
\end{proof}

\subsection{Proposition 1. }\label{appendix:prop1} 
Let us 
assume that
$ \forall  h\in\mathcal{ H}$, 
$\exists h'\in\mathcal{ H} $ s.t. $ \hat\ell(h'(\mathbf{w}),h(\mathbf{w}))=\phi'(\frac{p_s^w(\mathbf{w})}{p_t^w(\mathbf{w})}),\forall z\in\text{supp}(p_t^w(\mathbf{w})) $.
The maximizing solution of $d_{s,t}$ is  $\mathrm{D}_{{h}, {\mathcal{H}}}^{\phi}\left(P_{\mathrm{s}}^w\| P_{\mathrm{t}}^w\right)$, 
\ie $ \max_{h'\in\mathcal H}d_{s,t}=\mathrm{D}_{{h}, {\mathcal{H}}}^{\phi}\left(P_{\mathrm{s}}^w\| P_{\mathrm{t}}^w\right) $.

\begin{proof}
\begin{align*}
      d_{s, t}&=\mathbb{E}_{\mathbf{w}\sim P_{\mathrm{s}}^w}[\hat{\ell}\left({h}^{\prime}(\mathbf{w}), {h}(\mathbf{w})\right)] \\
      &-\mathbb{E}_{\mathbf{w} \sim P_{\mathrm{t}}^w}[(\phi^{*} \circ \hat{\ell})\left({h}^{\prime}(\mathbf{w}), {h}(\mathbf{w})\right)] \\
      &=\int p_{\mathrm{t}}^{\mathrm{w}}(\mathbf{w})[\frac{p_{\mathrm{s}}^{\mathrm{w}}(\mathbf{w})}{p_{\mathrm{t}}^{\mathrm{w}}(\mathbf{w})} \hat{\ell}\left({h}^{\prime}(\mathbf{w}), {h}(\mathbf{w})\right)\\
      &-\left(\phi^{*} \circ \hat{\ell}\right)\left({h}^{\prime}(\mathbf{w}), {h}(\mathbf{w})\right)] d \mathbf{w}.
\end{align*}

From the definition of Fenchel conjugate $\phi^*$, we have:
\begin{equation}\label{appedix:fench}
  x \in \partial \phi^{*}(t) \Longleftrightarrow \phi(x)+\phi^{*}(t)=x t \Longleftrightarrow \phi^{\prime}(x)=t.
\end{equation}

Let $t=\hat\ell(h(\mathbf{w}),h'(\mathbf{w})),\text{ and  }x=\frac{p_s^w(\mathbf{w})}{p_s^w(\mathbf{w})}$, then  $\forall \mathbf{w}\in\text{supp}(p_t^w(\mathbf{w}))$, plugging $\mathbf{w}$ into \eq{appedix:fench} :
\begin{align*}
    &\hat\ell\left({h}^{\prime}(\mathbf{w}), {h}(\mathbf{w})\right)=\phi^{\prime}\left(p_{\mathrm{s}}^{\mathrm{w}}(\mathbf{w}) / p_{\mathrm{t}}^{\mathrm{w}}(\mathbf{w})\right)\\
    &\Longleftrightarrow p_{\mathrm{s}}^{\mathrm{w}}(\mathbf{w}) / p_{\mathrm{t}}^{\mathrm{w}}(\mathbf{w})\in\partial \phi^*(\hat\ell(h'(\mathbf{w}),h(\mathbf{w}))).\\
   \label{pf-eq-17}
\end{align*}
Suppose $\hat\ell$ satisfies above condition, and the equality in \eq{eq-f-bound} holds.
Thus  
$ \max_{h'\in\mathcal H}d_{s,t}=D_{\phi}\left(P^w_{s} || P^w_{t}\right) $.
\end{proof}

\begin{table}[htbp]
	\caption{Common $f$-divergences and the corresponding convex function $\phi$.}
\label{tb-fdiv}
	\centering
\begin{tabular}{@{}lll@{}}
\toprule
Divergence  & Convex Function $\phi(x)$                                    & $\phi'(1)$     \\ \midrule
Kullback-Leibler        & $x\log x$                         & 1              \\
Jensen-Shannon        & $-(x+1)\log\frac{1+x}{2}+x\log x$              & 0              \\
Pearson $\chi^2$  & $(x-1)^2$      & 0              \\
Total Variation        & $\frac{1}{2}|x-1|$  & {[}$-0.5, 0.5${]} \\  \bottomrule
\end{tabular}
\end{table}
\subsection{Choice Pearson \texorpdfstring{$\chi^2$} -divergence }\label{appendix:pearson}
{Domain adaptation aims to minimize the discrepancy between two domains, which implies  $p_{\mathrm{s}}^{\mathrm{w}}(\mathbf{w}) / p_{\mathrm{t}}^{\mathrm{w}}(\mathbf{w})$ closing to  1 and the discrepancy closing to 0, for all $\mathbf{w}\in\text{supp}(p_t^w(\mathbf{w})) $. 
}
%
 
Maximizing  $d_{s,t}$ is  equivalent to calculating $D_{\phi}\left(P^w_{s} || P^w_{t}\right)$, when  $D_{\phi}\left(P^w_{s} || P^w_{t}\right)\to 0$ implies $\hat\ell\left(\hat{h}^{\prime}(\mathbf{w}), \hat{h}(\mathbf{w})\right)\to 0$ and Proposition 1 supposes that the convex function $\phi$ satisfies $\phi'(1)=0$. 
{
Recall that common examples of $f$-divergence and the convex functions in \tabref{tb-fdiv}.
We notice that the Pearson $\chi^2$ divergence satisfies $\phi'(1)=0$. }
Hence, Pearson $\chi^2$ divergence is chosen to calculate $d_{s,t}$ in this research.

Following  Proposition 1, the formulation for calculating the discrepancy $\mathrm{D}_{{h}, {\mathcal{H}}}^{\phi}\left(P_{\mathrm{s}}^w\| P_{\mathrm{t}}^w\right)$ is  as follows:
\begin{equation}\label{app:eq-minmax2}
\begin{aligned}
\mathrm{D}_{{h}, {\mathcal{H}}}^{\phi}\left(P_{\mathrm{s}}^w\| P_{\mathrm{t}}^w\right)&= \max_{h'\in\mathcal H}
\mathbb{E}_{\mathbf{w} \sim P_{\mathrm{s}}^w}\left[{\hat\ell}\left({h}^{\prime}, {h} \right)\right]  
\\&-\mathbb{E}_{\mathbf{w} \sim P_{\mathrm{t}}^w}\left[(\phi^{*} \circ {\hat\ell})\left({h}^{\prime} , {h} \right)\right] .
\end{aligned}
\end{equation}

{
Intuitively, regarding $G$ as the hypothesis function $h$ and $E$ maps images to latent space, the discrepancy is rewritten as:
\begin{equation}\label{app:eq-gan-adv}
\begin{aligned}
\max _{{G}^{\prime}}&\quad \mathbb{E}_{\mathbf x \sim P_{\mathrm{s}}}\left[\hat{\ell}\left({G}^{\prime} \circ E, {G} \circ E\right)\right]
\\&-\mathbb{E}_{\mathbf x \sim P_{\mathrm{t}}}\left[\left(\phi^{*} \circ \hat{\ell}\right)\left({G}^{\prime} \circ E, {G} \circ E\right)\right] .
\end{aligned}
\end{equation}
If we utilize an auxiliary generator $G'$ to estimate the divergence and L2 loss as $\hat \ell$ in~\eq{app:eq-gan-adv}, only optimizing the loss function in pixel level is insufficient to measure the difference between two images. In addition, the network structure of $G'$ is the same as the generator $G$~\cite{styleganv2}, leaving a larger amount of redundant model parameters.

To efficiently measure  the similarity between two images and decrease parameters, we project inverted images into the representation space of LPIPS \cite{LPIPS} to obtain the perceptual similarity between two images.
 Thus, the discrepancy is reformulated as:
\begin{equation}\label{app:eq-h-adv}
\begin{aligned}
\max _{\hat{H}} d_{s,t}&=  \mathbb{E}_{\mathbf x \sim P_{\mathrm{s}}}\left[\hat{\ell}\left(\hat{H} \circ G\circ E, H\circ {G} \circ E\right)\right] \\
&-\mathbb{E}_{\mathbf x \sim P_{\mathrm{t}}}\left[\left(\phi^{*} \circ \hat{\ell}\right)\left(\hat{H} \circ G\circ E, H\circ{G} \circ E\right)\right],
\end{aligned}
\end{equation}
where $\hat \ell$ refers to LPIPS loss which is interpreted as to minimize the discrepancy at the perceptual feature-level, and $\hat{H}$ indicates a network of the same structure as $H$ in LPIPS loss.
We summarize the proposed method in Algorithm~\ref{alg:gan_inv}.

}
\begin{algorithm}[!t]
    \caption{Pseudo-code of UDA-inversion}
    \label{alg:gan_inv}
    \textbf{Input}:  $\mathbf x_s$: image from source domain, $\mathbf{x}_t$: image from target domain, $G(\cdot)$:  pre-trained generator, $E(\cdot)$: encoder, $\hat H(\cdot)$: feature extracting network, $T$: the total number of iterations.
    \\
        \textbf{Initialization}: initialize $E^{(0)}$, $\hat H^{(0)}$.\\
        \textbf{For}{  $t = 1, 2, 3, \cdots, T$ \textbf{do}}{
            \\
             1.\quad Calculate $d_{s,t}$.\\
             2.\quad Update $\hat H^{(t+1)}$ by ascending  gradient: $\partial d_{s,t}/\partial\hat H^{(t)} $. \\
             3.\quad Compute $\mathcal{L}_s$ and  $d_{s,t}$ . \\
             4.\quad $\mathcal{L}=\mathcal{L}_s+\lambda_{uda}d_{s,t}$. \\
             5.\quad Update $E^{(t+1)}$ by  descending  gradient: $\partial \mathcal{L}/\partial E^{(t)} $.\\
         }
         \textbf{Output}: $E^{(T)}$ : the well trained encoder for GAN inversion. \\
    \end{algorithm}

\section{Experimental Details and Results}\label{supp:experimental details}

\subsection{Datasets and degradation operator}\label{appendix:deg}
Here, we provide additional details about datasets and degradation operations to obtain images in target domains.

FFHQ dataset contains 70,000 high-quality images. We use the first 35,000 images~\cite{stylegan} as the source domain, and the last 35,000 images  are degraded as the target domain.
In Stanford Cars dataset~\cite{iccvKrause}, we also use 50\% images from the training set as the source domain, and the rest of images are degraded as the target domain.
 Analogous to existing GAN inversion methods~\cite{idinvert,wang2021HFGI,e2style}, we resize the facial image resolution from $1024^2$ to $256^2$ and the car image resolution from $512\times 384$ to $256\times 192$ during training.

The target domain  consists of perturbed images (\ie LQ images) $\mathbf x_t$, which are degraded from HQ images $\mathbf x_s$: $ \mathbf x_t=\Psi (\mathbf x_s)$, where $\Psi$ denotes a degradation  operation.
The degradation operations contain rain layer, random mask, and down-sample, some of which are  utilized  in  existing methods\cite{psp,e2style}. 
 Particularly,  we utilized the `RainLayer' feature from the imgaug library~\cite{imgaug} with default parameters,  perturbing images to a  rain effect. 
 We employ the same free-form mask generation algorithm as GPEN~\cite{YangRX021} and the same bicubic down-sample with the factor of 2 as E2style~\cite{e2style}.

\subsection{Evaluation Metrics}\label{appendix:eval_m}
We quantitatively evaluate the inversion results from two aspects, i.e., inversion accuracy and image quality~\cite{xia2022gan}. 
Inversion accuracy is usually measured by PSNR (Peak Signal-to-Noise Ratio), SSIM (Structural Similarity)~\cite{ssim}, and MSE (Mean Squared Error).
Images generated based on GANs are usually evaluated by  FID (Fr\'echet inception distance)~\cite{fid} to assess the quality and diversity of generated distribution.

\begin{table}[!t]
\centering
\begin{tabular}{@{}cr|cr@{}}
\toprule
\multicolumn{2}{c|}{Face} & \multicolumn{2}{c}{Car} \\ \midrule
Parameters & Value & Parameters & Value \\ \midrule
$\lambda_1$ & 1 & $\lambda_1$ & 1 \\
$\lambda_2$ & 0.8 & $\lambda_2$ & 0.8 \\
$\lambda_3$ & 1 & $\lambda_3$ & 1 \\
$\lambda_4$ & 0.1 & $\lambda_4$ & 0.1 \\
$\lambda_{uda}$ & 1 & $\lambda_{uda}$ & 0.5 \\
Batch size & 8 & Batch size & 12 \\
Iteration & 200,000 & Iteration & 100,000 \\ \bottomrule
\end{tabular}
\caption{Hyper-parameters for our methods training on face and car dataset.}
\label{tab:hyperparameters}
\end{table}

\subsection{Implementation and Hyper-parameters}\label{appendix:implementation}

{
We use the pre-trained StyleGAN2~\cite{styleganv2} as the generator in experiments and adopt the encoder structure of HFGI~\cite{wang2021HFGI}, containing a  basic encoder $E_0$ and a consultation fusion branch to improve $E_0$ with higher fidelity.
The hyper-parameters of our encoder network are the same as HFGI.
In addition, we follow existing encoder-based methods ~\cite{wang2021HFGI,e2style} with Ranger optimizer~\cite{liu2019variance,zhang2019lookahead}  during training and set the initial learning rate to 0.0001.
Two popular editing methods, GANSpace~\cite{ganspace} and InterfaceGAN~\cite{interpGAN} are selected for semantic editing  to manipulate inverted images.
}

 All experiments are implemented on  NVIDIA TITAN RTX GPUs. The hyper-parameters for  the face dataset and car dataset are listed in \tabref{tab:hyperparameters}.






\end{appendices}

\end{document}